\documentclass[]{article}
\usepackage[utf8]{inputenc}
\usepackage{proceed2e}

\usepackage{natbib}
\usepackage{graphicx}
\usepackage{amsmath,amsfonts}
\usepackage{url}
\usepackage{amsthm}

\usepackage{times}

\title{On the inconsistency of $\ell_1$-penalised sparse precision matrix estimation}

\author{{\bf Otte Heinävaara}\\
  Helsinki Institute for Information Technology HIIT\\
  Department of Computer Science\\
  University of Helsinki\\
  \And
  {\bf Janne Leppä-aho}\\
  Helsinki Institute for Information Technology HIIT\\
  Department of Computer Science\\
  University of Helsinki\\
  \AND
  {\bf Jukka Corander}\\
  Helsinki Institute for Information Technology HIIT\\
  Department of Mathematics and Statistics\\
  University of Helsinki\\
  \And
  {\bf Antti Honkela}\\
  Helsinki Institute for Information Technology HIIT\\
  Department of Computer Science\\
  University of Helsinki}

\newcommand\independent{\protect\mathpalette{\protect\independenT}{\perp}}
\def\independenT#1#2{\mathrel{\rlap{$#1#2$}\mkern2mu{#1#2}}}

\newcommand{\R}{\mathbb{R}}
\newcommand{\vx}{\mathbf{x}}
\newcommand{\vy}{\mathbf{y}}
\newcommand{\vA}{\mathbf{A}}
\newcommand{\vC}{\mathbf{C}}
\newcommand{\vLambda}{\boldsymbol{\Lambda}}
\newcommand{\vSigma}{\boldsymbol{\Sigma}}
\newcommand{\vOmega}{\boldsymbol{\Omega}}
\newcommand{\bBeta}{\boldsymbol{\beta}}
\theoremstyle{plain}
\newtheorem{assumption}{Assumption}

\begin{document}

\maketitle

\begin{abstract}
  Various $\ell_1$-penalised estimation methods such as graphical
  lasso and CLIME are widely used for sparse precision matrix
  estimation.  Many of these methods have been shown to be consistent
  under various quantitative assumptions about the underlying
  true covariance matrix.  Intuitively, these conditions are related to
  situations where the penalty term will dominate the optimisation.
  In this paper, we explore the consistency of $\ell_1$-based methods
  for a class of sparse latent
  variable -like models, which are strongly motivated by several types of applications. We show that all $\ell_1$-based methods fail
  dramatically for models with nearly linear dependencies between the
  variables. We also study the consistency on models derived from
  real gene expression data and note that the assumptions needed for
  consistency never hold even for modest sized gene networks and
  $\ell_1$-based methods also become unreliable in practice for larger
  networks.
\end{abstract}

\section{INTRODUCTION}

Estimating the sparse precision matrix, i.e.\ the inverse covariance
matrix, from data is a very widely used method for exploring the dependence
structure of continuous variables.  The motivation for the approach stems from the
fact that for a Gaussian Markov random field model, zeros in the
precision matrix translate exactly to absent edges in the corresponding
undirected Gaussian graphical model, thus being informative about the
marginal and conditional independence relationships among the variables.

The full $p$-dimensional covariance matrix contains $p(p+1)/2$ parameters,
making its accurate estimation from limited data difficult.  Additionally, the structure learning requires the inverse of the
covariance, and matrix inversion is in general a very fragile
operation.  To make the problem tractable, some form of regularisation
is typically needed.  Direct optimisation of the sparse structure would easily
lead to very difficult combinatorial optimisation problems.  To avoid
these computational difficulties, several convex
$\ell_1$-penalty-based approaches have been proposed.  Popular
examples include $\ell_1$-penalised maximum likelihood estimation
\citep{Meinshausen2006}, which also forms the basis for the highly
popular graphical lasso (glasso) algorithm \citep{Friedman2008}.
$\ell_1$ regularisation has also been used for example in a non-probabilistic
alternative with linear-programming-based constrained $\ell_1$
minimisation (CLIME) algorithm of \citet{Cai2011}.

At the heart of the optimisation problems considered by all these
methods is a term depending on the $\ell_1$ norm of the estimated
precision matrix.  $\ell_1$-penalisation-based approaches such as
lasso are popular for sparse regression, but they have a known
weakness: in addition to promoting sparsity they also push true
non-zero elements toward zero~\citep{Zhao2006}.  In the context of
precision matrix estimation this effect would be expected to be
especially strong when some elements of the precision matrix are large,
which happens for scaled covariance matrices when the covariance matrix becomes ill-conditioned.
This phenomenon occurs frequently under the circumstances where some of the variables are nearly linearly dependent.

In this paper we demonstrate a drastic failure of the $\ell_1$
penalised sparse covariance estimation methods for a class of models
that have a linear latent variable structure where some variables
depend linearly on others.  For such models even in the limit of
infinite data, popular $\ell_1$ penalised methods cannot yield
results that are significantly better than based on random guessing on any
setting of the regularisation parameter.  Yet these models have a very
clear sparse structure that becomes obvious from the empirical
precision matrix with an increasing $n$.

Given the huge popularity and success of linear models in modelling
data, structures like the one considered in our work are natural for various real world
data sets.  Motivated by our discovery, we also explore the
inconsistency of $\ell_1$ penalised methods on models derived from
real gene expression data and find them poorly suited for such applications.

\section{STRUCTURE LEARNING OF GAUSSIAN GRAPHICAL MODELS}
\subsection{BACKGROUND}
We start with a quick recap on the basics of Gaussian graphical models in order to formulate the problem of structure learning. For a more comprehensive treatment of the subject, we refer to (\citealt{WHITTAKER1990}; \citealt{LAURITZEN1996}). Let $\mathbf{X} = (X_1, \ldots , X_p)'$ denote a random vector following a multivariate normal distribution with zero mean and a covariance matrix $\mathbf{\Sigma}$, $\mathbf{X} \sim N_p(\mathbf{0}, \mathbf{\Sigma}) $. Let $G = (V,E)$ be an undirected graph, where the $V = \{ 1, \ldots , p \}$ is the set of nodes and $E \subset V\times V$ stands for the set of edges. The nodes in the graph represent the random variables in the vector $\mathbf{X}$ and absences of the edges in the graph correspond conditional independence assertions between these variables. More in detail, we have that $(i,j)\not\in E$ and $(j,i)\not\in E $ if and only if $X_i$ is conditionally independent of $X_j$ given the remaining variables in $\mathbf{X}$.

In the multivariate normal setting, there is a one-to-one correspondence between the missing edges in the graph and the off-diagonal zeros of the precision matrix $\mathbf{\Omega} = \mathbf{\Sigma}^{-1}$, that is, $\omega_{ij} = 0 \Leftrightarrow X_i \independent X_j \ | \mathbf{X} \setminus \{X_i, X_j \}$ (see, for instance, \citealt{LAURITZEN1996}, p.\ 129). Given an undirected graph $G$, a Gaussian graphical model is defined as the collection of  multivariate normal distributions for $\mathbf{X}$ satisfying the conditional independence assertions implied by the graph $G$.

Assume we have a complete (no missing observations) i.i.d. sample $\mathbf{x} = (\mathbf{x}_1,\ldots , \mathbf{x_n})$ from the distribution $N_p(\mathbf{0}, \mathbf{\Sigma})$. Based on the sample $\mathbf{x}$, our goal in structure learning is to find the graph $G$, or equivalently, learn the zero-pattern of $\mathbf{\Omega}$. The usual assumption is that the underlying graph is sparse. A naive estimate for $\mathbf{\Omega}$ by inverting the sample covariance matrix is practically never truly sparse for any real data. Furthermore, if $n < p$ the sample covariance matrix is rank-deficient and thus not even invertible.

One common approach to overcome these problems is to impose an additional $\ell _1$-penalty on the elements of $\mathbf{\Omega}$ when estimating it. This kind of regularisation effectively forces some of the elements of $\mathbf{\Omega}$ to zero, thus resulting in sparse solutions. In the context of regression models, this method applied on the regression coefficients goes by the name of \textit{lasso} \citep{Tibshirani1996}. There exists a wide variety of methods making use of $\ell _1$-regularisation in the setting of Gaussian graphical model structure learning (\citealt{Yuan07}; \citealt{Meinshausen2006}; \citealt{Banerjee2008}; \citealt{Friedman2008}; \citealt{Peng09}; \citealt{Cai2011}; \citealt{Hsieh2014}).

\subsection{$\ell_1$-REGULARISED METHODS}

In this section we provide a brief review of selected examples of
different types of $\ell_1$-penalised methods.

\subsubsection{Glasso}
We begin with the widely used graphical lasso-algorithm (glasso) by \cite{Friedman2008}. Glasso-method maximises an objective function consisting of the Gaussian log-likelihood and an $\ell _ 1$-penalty:
\begin{equation}\label{eq:glassoObj}
\log\det(\mathbf{\Omega}) - \textnormal{trace}(\mathbf{\Omega S}) - \lambda ||\mathbf{\Omega}||_1,
\end{equation}where $\mathbf{S}$ denotes the sample covariance matrix and $\lambda > 0$ is the regularisation parameter controlling the sparsity of the solution. The $\ell_1$ penalty, $||\vOmega||_1= \sum_{i,j}|\omega_{ij}|$, is applied on all the elements of $\vOmega$, but the variant where the diagonal elements are omitted is also common. The objective function (\ref{eq:glassoObj}) is maximised over all positive definite matrices $\mathbf{\Omega}$ and the optimisation is carried out in practice using a block-wise coordinate descent.

\subsubsection{CLIME}
\cite{Cai2011} approach the problem of sparse precision matrix estimation from a slightly different perspective. Their CLIME-method (Constrained $\ell_1$-minimisation for Inverse Matrix Estimation) seeks matrices $\vOmega$ with a minimal $\ell_1$-norm under the following constraint
\begin{equation}\label{eq:CLIME}
|\mathbf{S}\vOmega - \mathbf{I}|_{\infty} \leq \lambda,
\end{equation} where $\lambda$ is the tuning parameter and $|\mathbf{A}|_\infty = \max_{i,j} |a_{ij}|$ is the element-wise maximum. The optimisation problem $\min_{\vOmega} ||\vOmega||_1$ subject to the constraint (\ref{eq:CLIME}) does not explicitly force the solution to be symmetric, which is resolved by picking from estimated values $\omega_{ij}$ and $\omega_{ji}$ the one with a smaller magnitude into the final solution. In practice, the optimisation problem is decomposed over variables into $p$ sub-problems which are then efficiently solved using linear programming.

\subsubsection{SCIO}
\cite{Liu2015} introduced recently a method called Sparse Column-wise Inverse Operator (SCIO). The SCIO-method decomposes the estimation of $\vOmega$ into a following smaller problems
$$
\min_{\bBeta_i \in \R^{p}} \left\{ \frac{1}{2} \bBeta_i^T \mathbf{S} \bBeta_i - \textbf{e}_i^T \bBeta_i + \lambda ||\bBeta_i ||_1 \right\},
$$where $\mathbf{S}$ and $\lambda$ are defined as before and $\textbf{e}_i$ is an $i$:th standard unit vector. The regularisation parameter $\lambda$ can in general vary with $i$ but this is omitted in our notation. The solutions $\hat{\bBeta}_i$ form the columns for the estimate of $\vOmega$. Also for SCIO, the symmetry of the resulting precision matrix must be forced, and this is done as described in the case of CLIME. 

\subsection{ALTERNATIVE METHODS}

\subsubsection{The naive approach}
In addition to the above-mentioned $\ell_1$-penalised methods, we consider two alternative approaches. In a "naive" approach, we simply take the sample covariance matrix, invert it, and then threshold the resulting matrix to obtain a sparse estimate for the precision matrix. The threshold value is chosen using the ground truth graph so that the naive estimator will have as many non-zero entries as there are edges in the true graph. Setting the threshold value according to the ground truth is of course unrealistic, however, it is nevertheless interesting to compare the accuracy of this simple procedure to the performance of the more refined $\ell_1$-methods, when also their tuning parameters are chosen in a similar fashion.

\subsubsection{FMPL}
Lastly, we consider a Bayesian approach which is based on finding a graph with a highest fractional marginal pseudo-likelihood (FMPL) by \cite{FMPL}. The fractional marginal pseudo-likelihood is an approximation of the marginal likelihood and it has been shown to be a consistent scoring function in the sense that the true graph maximises it as the sample size tends to infinity, under the assumption that data are generated from a multivariate normal distribution. The FMPL-score decomposes over variables and in practice, the method identifies optimal Markov blankets for each of the variables, which are then combined into a proper undirected graph using any of the three different schemes commonly employed in graphical model learning: OR, AND and HC.
 
\subsection{MODEL SELECTION CONSISTENCY}
The assumptions required for a consistent model selection with an $\ell _ 1$-penalised Gaussian log-likelihood have been studied, for instance, in \cite{Ravikumar2011}. The authors provide a number of conditions in the multivariate normal model that are sufficient for the recovery of the zero pattern of the true precision matrix $\vOmega^*$ with a high probability when the sample size is large. For our purposes, the most relevant condition is the following:
\begin{assumption}\label{assumption1} There exists $\alpha \in (0, 1],$ such that 
  \begin{equation}
    \label{eq:assumption1}
    \gamma := ||\Gamma_{S^CS}{(\Gamma_{SS})}^{-1}||_\infty \leq 1 - \alpha.
  \end{equation}
\end{assumption}
Here $S \subset V \times V$ is a set defining the support of $\vOmega^*$, that is, the non-zero elements of $\vOmega^*$ (diagonal and the elements corresponding to the edges in the graphical model) and  $S^C$ refers to the complement of $S$ in $V\times V$. The $\Gamma$ term is defined via Kronecker product $\otimes$ as $\Gamma = (\vOmega^*)^{-1} \otimes (\vOmega^*)^{-1}\in\R^{p^2\times p^2}$ and $\Gamma_{AB}$ refers to the specific rows and columns of $\Gamma$ indexed by $A\subset V \times V$ and $B\subset V \times V$, respectively. The norm in the equation is defined as $||A||_\infty = \max_j \sum_i |a_{ij}|$.

The above result applies to glasso. However, a quite similar result was presented for SCIO in \cite{Liu2015}:   
\begin{assumption}\label{assumption2} There exists $\alpha \in (0, 1),$ such that 
$$
\max_{1 \leq i \leq p }||\vSigma^*_{\textbf{s}_i^C\textbf{s}_i}{(\vSigma^*_{\textbf{s}_i\textbf{s}_i})}^{-1}||_\infty \leq 1 - \alpha.
$$ 
\end{assumption}
Here $\vSigma^* = (\vOmega^*)^{-1}$ and $\textbf{s}_{i} = \{ j \in \{1,\ldots ,  p\} \ |  \ (\vOmega^*)_{ij} \neq 0 \}$. Assumption \ref{assumption2} under the multivariate normality guarantees that the support of $\vOmega^*$ is recovered by SCIO with a high probability as the sample size gets large.

\section{LATENT VARIABLE LIKE MODELS INDUCE INCONSISTENCY WITH $\ell_1$-PENALISATION}
\label{sec:latent-variable-like}

Methods for sparse precision matrix estimation generally depend on an
objective function (such as log-likelihood) and a penalty function or
regulariser, which in a Bayesian setting is usually represented by the
prior.  The ideal penalty function for many problems would be the
$\ell_0$ ``norm'' counting the number of non-zero elements: $||x||_0 =
\#\{i | x_i \neq 0\}$.  This $\ell_0$ function is not a proper norm,
but it provides a very intuitive notion of sparsity.  The main problem
with its use is computational: using $\ell_0$-penalisation leads to
very difficult non-convex combinatorial optimisation problems.  The
most common approach to avoid the computational challenges is to use
$\ell_1$-penalisation as a convex relaxation of $\ell_0$.  As mentioned
above this works well in many cases but it comes with a price, since in addition
to providing the sparsity, $\ell_1$ also regularises large non-zero values.
Depending on the problem, as we demonstrate here, this effect can be substantial and may
cause $\ell_1$-regularised methods to return totally meaningless
results.

Intuitively, $\ell_1$-regularised methods are expected to fail when
some elements of the true precision matrix become so large that their
contribution to the penalty completely overwhelms the other parts of
the objective and the penalty.  One example where this happens is when
some set of variables depends linearly on another set of variables.
In such situation the covariance matrix can become ill-conditioned and
the elements of its inverse, the precision matrix, grow.  One example
of when this happens is models with a linear latent variable
structure.

\begin{figure*}[htb]
  \centering
  \includegraphics{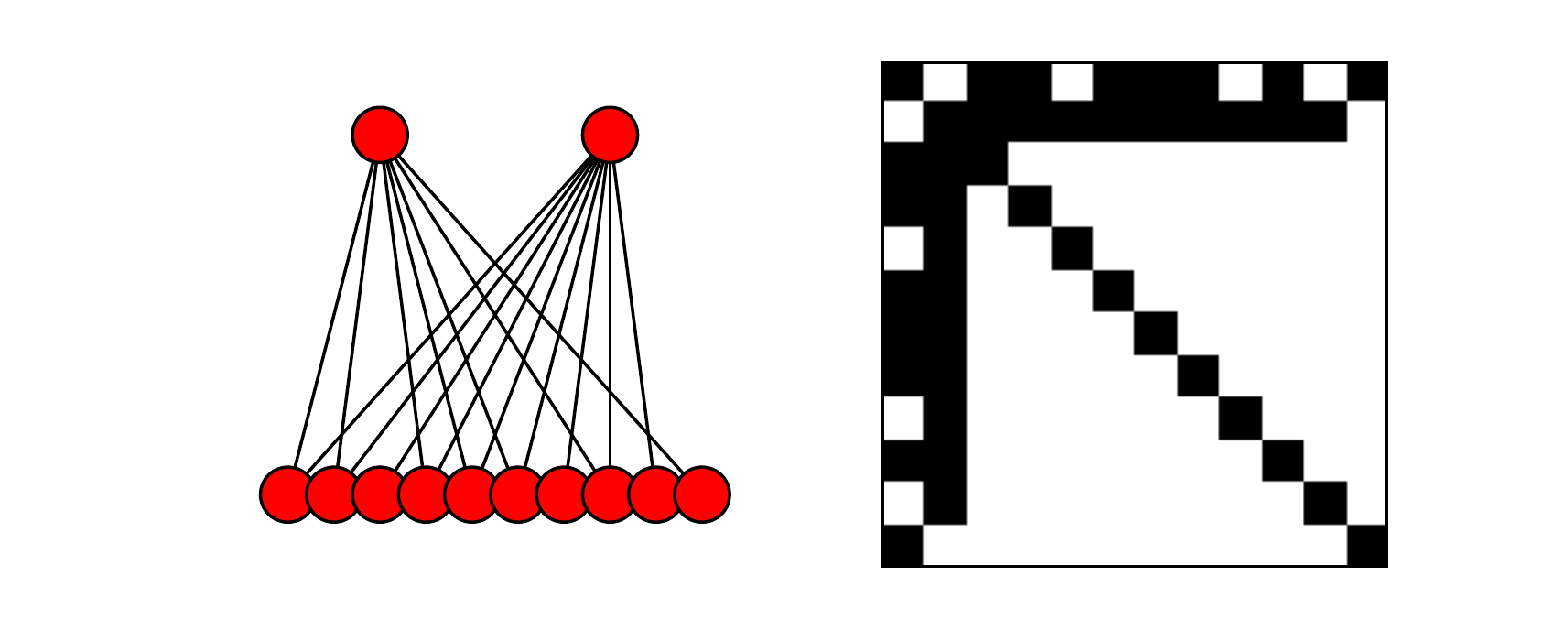}
  \caption{Left: Graphical representation of a latent variable model
    as an undirected graphical model for a case with somewhat sparse
    $\vA$. Right: The adjacency matrix of the graph showing the sparse
    pattern of non-zero elements in the corresponding precision
    matrix.}
  \label{fig:latent_variable_structure}
\end{figure*}

Let us consider a model for $\vx \in \R^{d_1}, \vy \in \R^{d_2}$,
where $\vy = \vA \vx + \epsilon$.  The graphical structure of the
model and the corresponding precision matrix structure are illustrated
in Fig.~\ref{fig:latent_variable_structure}.
Assuming $\vx \sim \mathcal{N}(0,
\sigma_x^2 I), \epsilon \sim \mathcal{N}(0, \sigma_{\epsilon}^2 I)$, the
covariance of the concatenated vectors $(\vx^T, \vy^T)^T$ is
given by the block matrix
\begin{equation}
  \label{eq:latent_model_cov}
  \mathrm{Cov}((\vx^T, \vy^T)^T) = \mathbf{C} = \sigma_x^2
  \begin{pmatrix}
    I & \vA^T \\
    \vA & \vA \vA^T + \sigma_{\epsilon}^2 I
  \end{pmatrix}.
\end{equation}
The covariance matrix has an analytical block matrix inverse \citep{Lu2002}
\begin{equation}
  \label{eq:latent_model_prec}
  \mathbf{C}^{-1} = \sigma_x^{-2}
  \begin{pmatrix}
    I + \sigma_{\epsilon}^{-2} \vA^T \vA & -\sigma_{\epsilon}^{-2} \vA^T \\
    -\sigma_{\epsilon}^{-2} \vA & \sigma_{\epsilon}^{-2} I
  \end{pmatrix}.
\end{equation}
This precision matrix recapitulates the conditional independence result
for Gaussian Markov random fields: the lower right block is diagonal
because the variables in $\vy$ are conditionally independent of each
other given $\vx$.  The matrix is clearly sparse, so we would
intuitively assume sparse precision matrix estimation methods should
be able to recover it.  The non-zero elements do, however, depend on
$\sigma_{\epsilon}^{-2}$ which can make them very large if the noise
$\sigma_{\epsilon}^{2}$ is small.

It is possible to evaluate and bound the different terms of
Eq.~(\ref{eq:glassoObj}) evaluated at the ground truth for these
models:
\begin{align*}
  \log\det(\mathbf{C}^{-1}) &= -d_2 \log \sigma_{\epsilon}^2 \\
  -\textnormal{trace}(\mathbf{C} \mathbf{C}^{-1}) &= -(d_1 + d_2) \\
  -\lambda ||\mathbf{C}^{-1}||_1 &< -\lambda \sigma_{\epsilon}^{-2} (d_2 + 2 ||\vA||_1).
\end{align*}
The magnitude of the last penalty term clearly grows very quickly as
$\sigma_{\epsilon}^{2}$ decreases.  Clearly the magnitude of the two
first log-likelihood terms grows much more slowly as they only depend
on $\log \sigma_{\epsilon}^{2}$.  Thus the total value of
Eq.~(\ref{eq:glassoObj}) decreases without bound as
$\sigma_{\epsilon}^{2}$ decreases.

Forgetting the ground truth, it is easy to see that one can construct
an estimate $\mathbf{\Omega}$ for which the objective remains bounded.
If we assume all values of $\mathbf{C}$ to be $\le 1$ (after normalisation),
$$\textnormal{trace}(\mathbf{C} \mathbf{\Omega}) \le ||\mathbf{\Omega}||_1.$$
As the other terms only depend on $\mathbf{\Omega}$ it is easy to
choose $\mathbf{\Omega}$ so that they remain bounded.  The estimate
$\mathbf{\Omega}$ that yields these values will in many cases not have
anything to do with $\mathbf{C}^{-1}$, as seen in the experiments
below.

\section{EXPERIMENTS}

We tested the performance of glasso, SCIO and CLIME as well as FMPL
using the model structure introduced in
Sec.~\ref{sec:latent-variable-like}.
The performance of the methods was investigated by varying the noise
variance $\sigma_{\epsilon}^{2}$, and the sample size $n$.
The model matrix $\vA$ was created as a $(d_{2},d_{1})$-array of independent normal random variables with mean $0$ and variance $1$.
The majority of the tests were run using input dimensionality
$d_1 = 2$, output dimensionality $d_2 = 10$ and noise variance
$\sigma_{\epsilon}^2 = 0.1^2$ but we also tested
varying these settings.
For each individual choice of noise and sample size, $k = 50$ different matrices $\vA$ were generated and the results were averaged.

Generating $n$ samples using model described, data were normalised and
analysed using the five different methods. We calibrated the methods in a way that number of
edges in the resulting graph would match the true number. Similarly,
we thresholded the naive method by taking inverse matrix directly to
output the correct number of edges.  The FMPL method has no direct tuning
parameters so we used its OR mode results as such.  Similar tuning is
not possible in a real problem where the true number of edges is now
known.  The tuning represents the best possible results the methods could
obtain with an oracle that provides an optimal regularisation
parameter.

We evaluated the results using the Hamming distance between the ground
truth and the inferred sparsity pattern, i.e.\ the number of incorrect
edges and non-edges which were treated symmetrically.  For methods
returning the correct number of edges, this value is directly related
to the precision $pr$ through
\begin{equation*}
  d_{\text{Hamming}} = 2 (1 - pr) N_{\text{true positives}}
\end{equation*}
or conversely
\begin{equation*}
  pr = 1 - \frac{d_{\text{Hamming}}}{2 N_{\text{true positives}}}.
\end{equation*}
We will nevertheless use the Hamming distance as it enables fair
comparison with FMPL that sometimes returns a different number of
edges.

\begin{figure}[htb]
  \centering
  \includegraphics[width=\columnwidth]{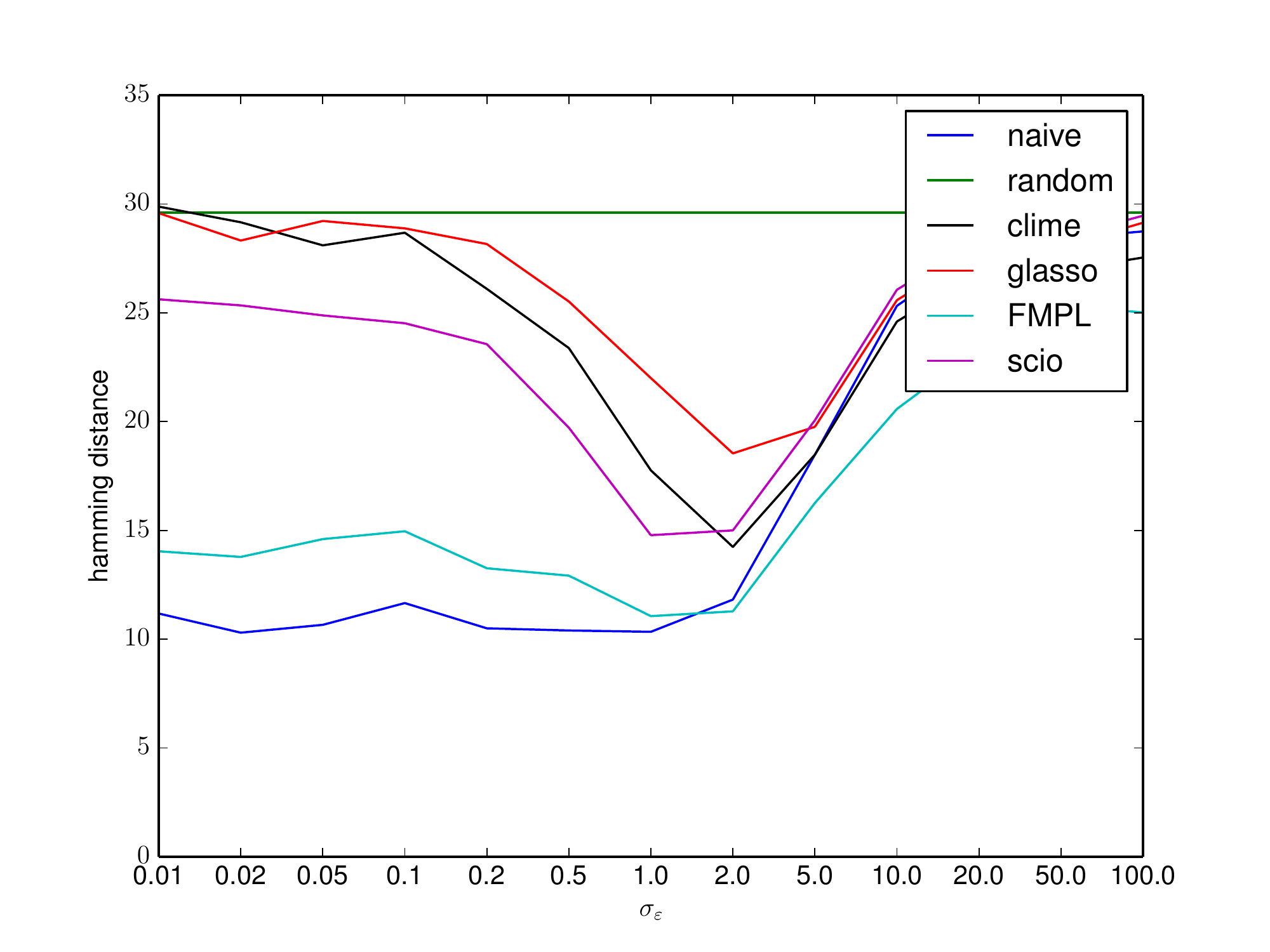}
  \caption{Performances of different methods on latent variable like
    model with 100 samples. (Lower values are better.)}
  \label{fig:noise_n100}
\end{figure}

\begin{figure}[htb]
  \centering
  \includegraphics[width=\columnwidth]{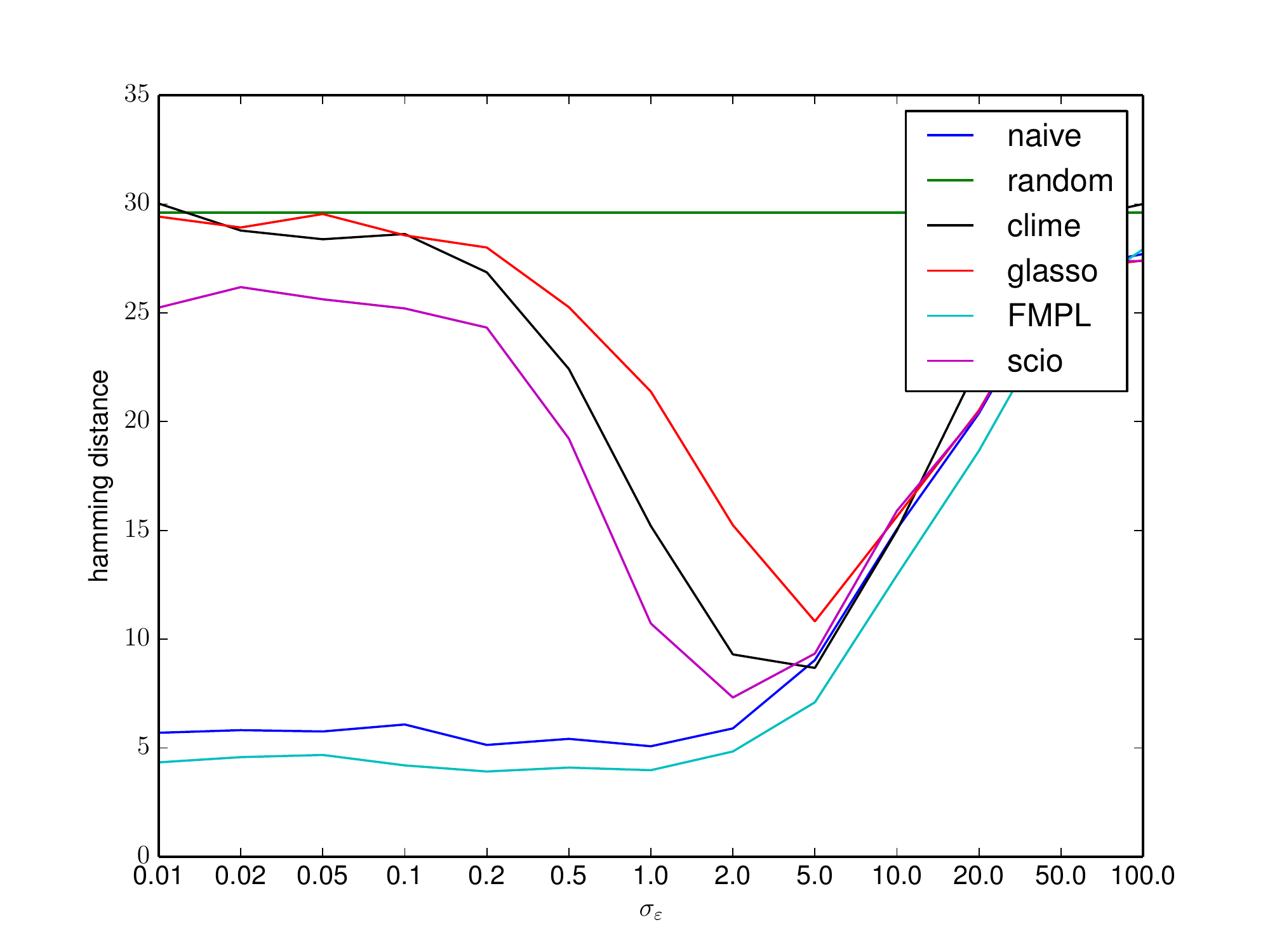}
  \caption{Performances of different methods on latent variable like
    model with 1000 samples. (Lower values are better.)}
  \label{fig:noise_n1000}
\end{figure}

Figs.~\ref{fig:noise_n100} and \ref{fig:noise_n1000} show the Hamming
distance obtained by the different methods as a function of the noise
level when using 100 and 1000 samples, respectively.  The results show
that especially for low but also for high noise levels, the
$\ell_1$-based methods all perform very poorly with especially glasso
and CLIME performing very close to random guessing level for low noise
levels $\sigma_{\epsilon} \le 0.1$.  The naive inverse and FMPL work
much better up to moderate noise levels of
$\sigma_{\epsilon} \approx 2$ after which the noise starts to dominate
the signal and the performance of all methods starts to drop.  SCIO is
a little better than the other $\ell_1$-based methods but clearly
worse than FMPL and naive in the low noise regime.

\begin{figure}[htb]
  \centering
  \includegraphics[width=\columnwidth]{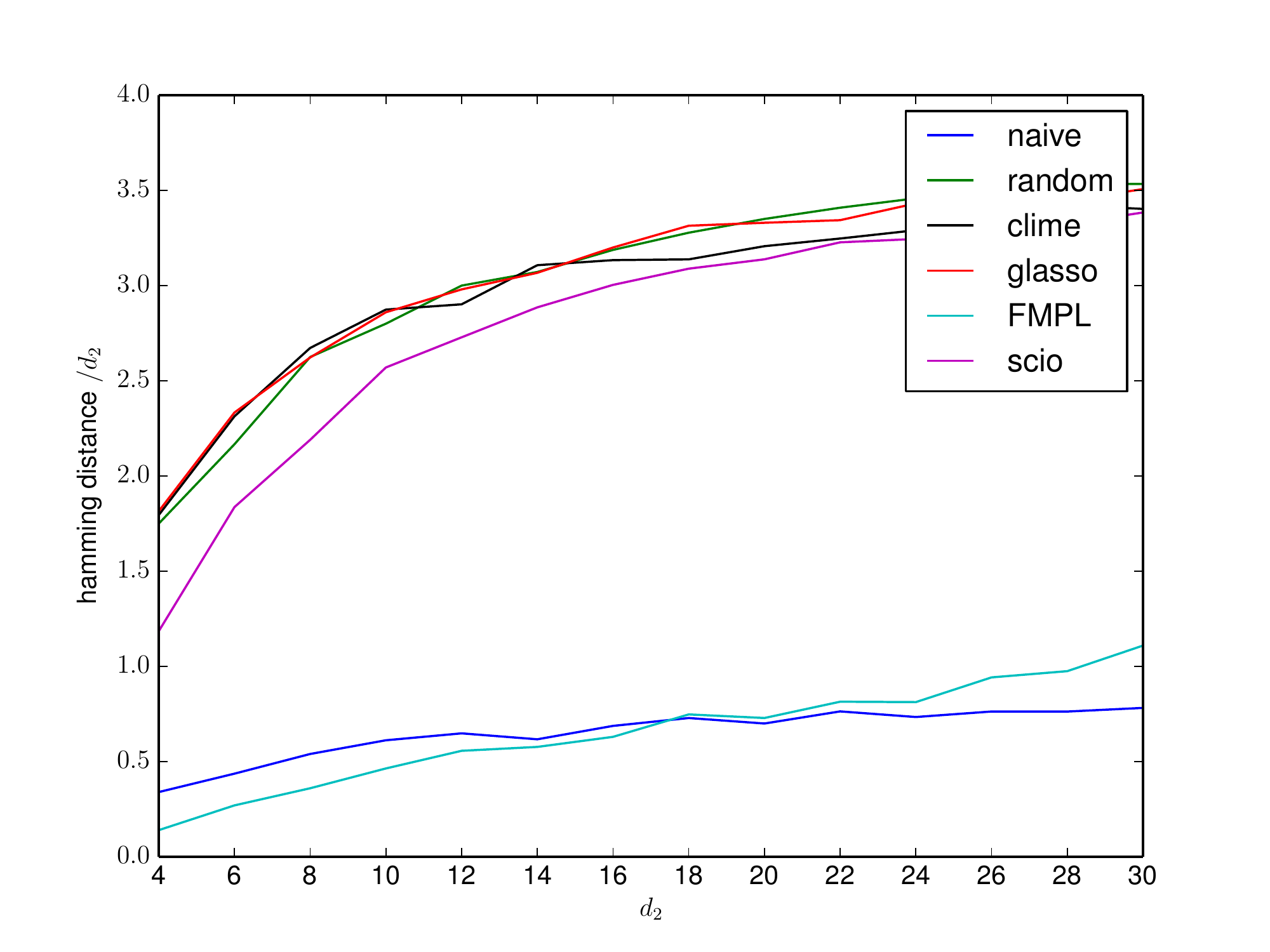}
  \caption{Performances of different methods on latent variable like
    model with varying output dimensionality. (Lower values are better.)}
  \label{fig:outdim_n1000}
\end{figure}

\begin{figure}[htb]
  \centering
  \includegraphics[width=\columnwidth]{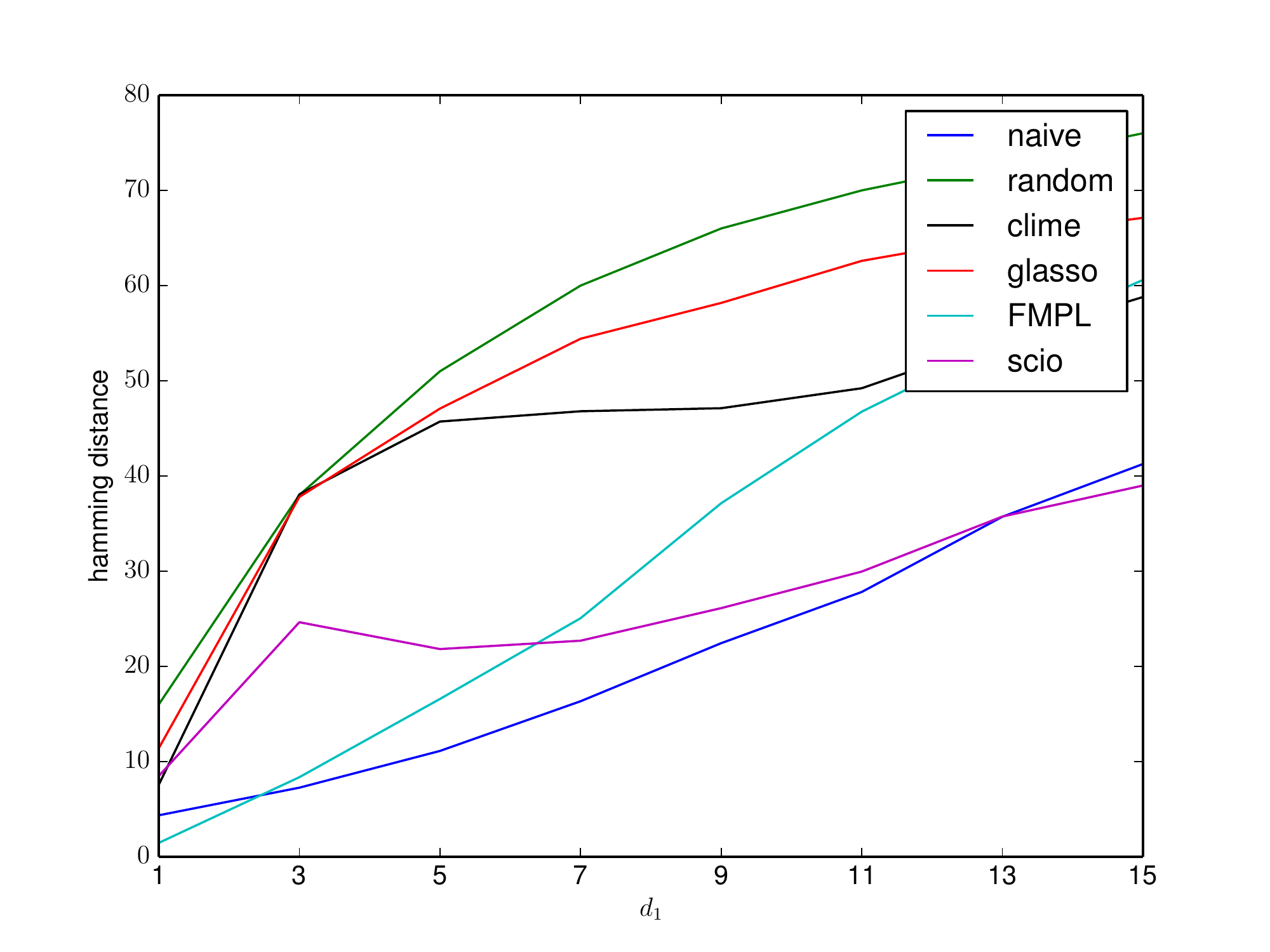}
  \caption{Performances of different methods on latent variable like
    model with varying input dimensionality. (Lower values are better.)}
  \label{fig:indim_n1000}
\end{figure}

Fig.~\ref{fig:outdim_n1000} shows the results when changing the output
dimensionality $d_2$ from 10.  The results show that the performance
of all $\ell_1$-based methods is very poor across all $d_2$.  Glasso
performance is close to random guessing level across the entire range
considered, while CLIME is slightly better for $d_2 \ge 18$ and SCIO
slightly better across the entire range.  Both FMPL and naive are
significantly better than any of the $\ell_1$-based methods.

Fig.~\ref{fig:indim_n1000} shows the corresponding result when
changing the input dimensionality $d_1$.  The results are now quite
different as all methods are better than random especially for larger
values.  SCIO still outperforms CLIME which outperforms glasso.  FMPL
is really accurate for small $d_1$ but degrades for larger $d_1$ while
the naive method is the most accurate in almost all cases.

\begin{figure}[htb]
  \centering
  \includegraphics[width=\columnwidth]{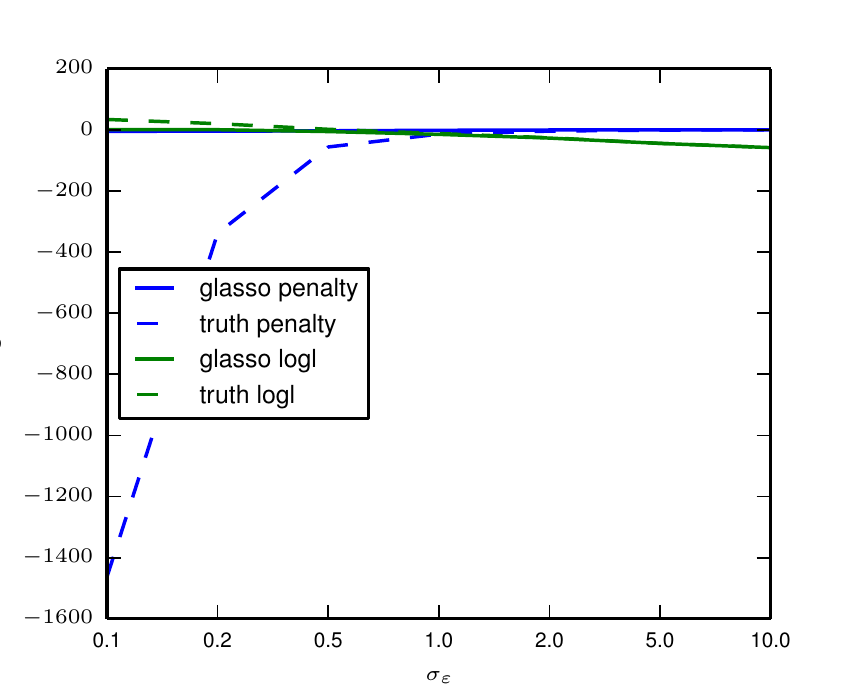}
  \caption{Contributions of the different terms of the glasso
    objective (\ref{eq:glassoObj}) for latent variable like model with
    1000 samples.  The green curves show the contributions of the
    first two terms of Eq.~(\ref{eq:glassoObj}) and the blue curves
    show the contributions of the last penalty term.  Solid lines
    show the result of the glasso optimal solution while dashed lines
    show the result for the true solution.}
  \label{fig:obj_n100}
\end{figure}

To further illustrate the behaviour of glasso on these examples,
Fig.~\ref{fig:obj_n100} shows the contributions of the different parts
of the glasso objective function (\ref{eq:glassoObj}) as a function of
the noise level both for the true solution (``truth'') as well as the
glasso solution.  The results show that for low noise levels the
penalty incurred by the true solution becomes massive.  The glasso
solution has a much lower log-likelihood (``logl'') than ground truth
but this is amply compensated by the significantly smaller penalty.
As the noise increases, the penalty of the true solution decreases
and the glasso solution converges to similar values.

\subsection{NECESSITY OF ASSUMPTION 1}

\begin{figure}[htb]
  \centering
  \includegraphics[width=\columnwidth]{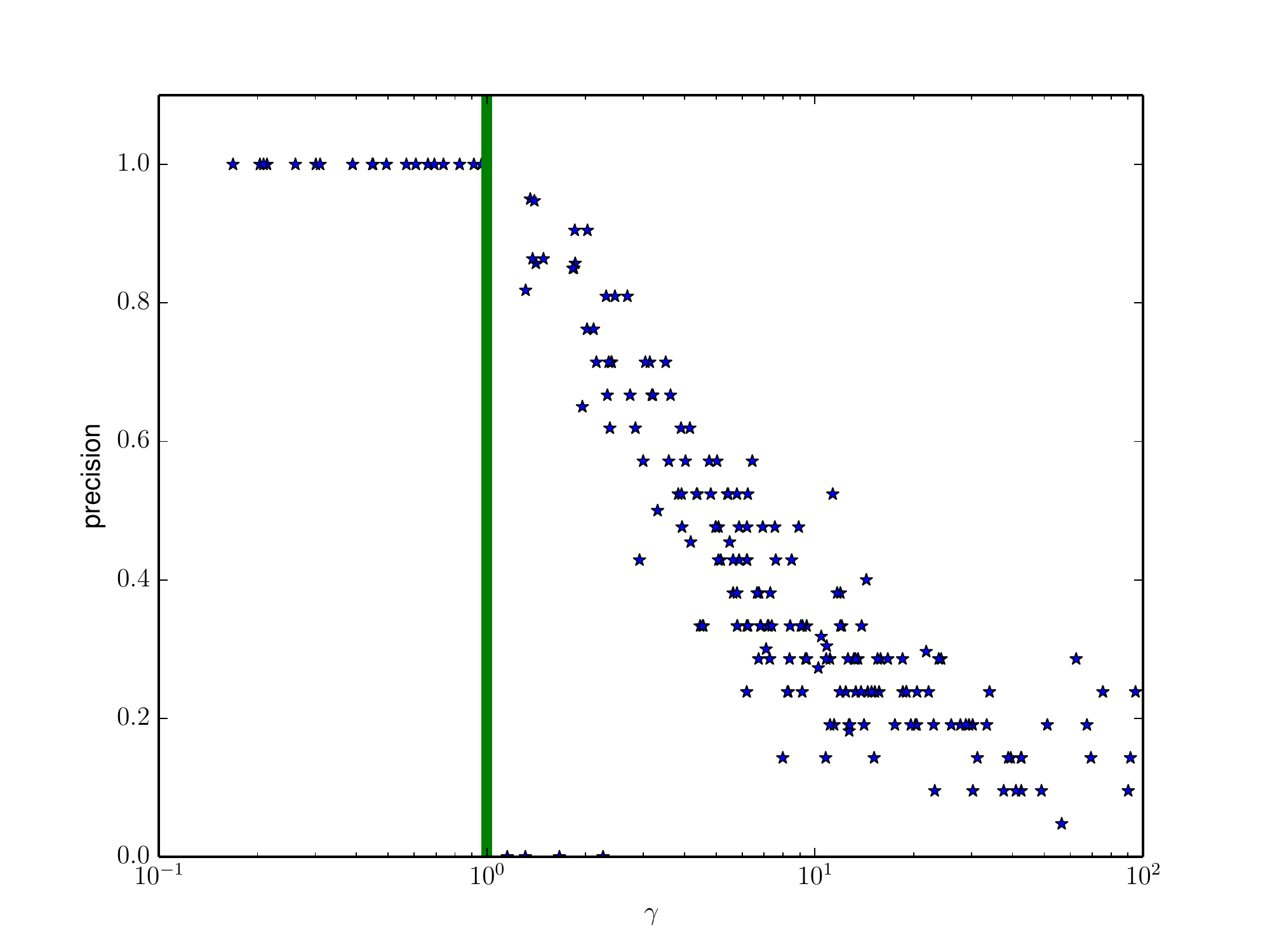}
  \caption{Precision of glasso on infinite data as a function of the
    norm $\gamma$ of Assumption 1 and Eq.~(\ref{eq:assumption1}).
    Values to the left of the green vertical line satisfy this
    condition while values to the right violate it. (Higher values are better.)}
  \label{fig:latent_assumption1}
\end{figure}

It can be checked that
the norm $\gamma$ in Assumption 1 and Eq.~(\ref{eq:assumption1}) for
latent-variable-like models depends on the scale of $\vA$.  We took
advantage of this by creating examples with different values of
$\gamma$ and testing the precision of glasso using the true covariance
which corresponds to infinite data limit.  The results of this
experiment are shown in Fig.~\ref{fig:latent_assumption1}.  The
results verify that glasso consistently yields perfect results when
$\gamma < 1$ which is a part of the sufficient conditions for
consistency of glasso.  As $\gamma$ grows and the sufficient
conditions are no longer satisfied, it is clearly seen that the accuracy of
glasso starts to deteriorate rapidly.  This suggests that the
sufficient condition of Assumption 1 is in practice also necessary to
ensure consistence.

\section{INCONSISTENCY FOR MODELS OF REAL GENE EXPRESSION DATA}

We tested how often the problems presented above appear in real data
using the ``TCGA breast invasive carcinoma (BRCA) gene expression by
RNAseq (IlluminaHiSeq)'' data set \citep{TCGA2012} downloaded from
\url{https://genome-cancer.ucsc.edu/proj/site/hgHeatmap/}.  The data
set contains gene expression measurements for 20530 genes for $n=1215$
samples. After removing genes with a constant expression across all
samples there are $p=20252$ genes remaining.

In order to test the methods we randomly sampled subsets of $d$ genes
and considered the correlation matrix $\vC_0$ over that subset.  We
generated sparse models with known ground truth by computing the
corresponding precision matrix $\vLambda_0$ from the empirical correlation matrix,
setting elements with absolute values below chosen cutoff $\delta=0.1$
to 0 to obtain
\begin{equation}
  \label{eq:gene_expression_prec}
  \vLambda_{ij} =
  \begin{cases}
    (\vLambda_0)_{ij} \quad &\text{if } |(\vLambda_0)_{ij}| > \delta \\
    0 \quad &\text{otherwise}
  \end{cases}
\end{equation}
and the testing covariance matrix $\vC = \vLambda^{-1}$. The cutoff
lead to networks that were sparse with on average 60\% zeros in the
precision matrix.

\begin{figure}[htb]
  \centering
  \includegraphics[width=\columnwidth]{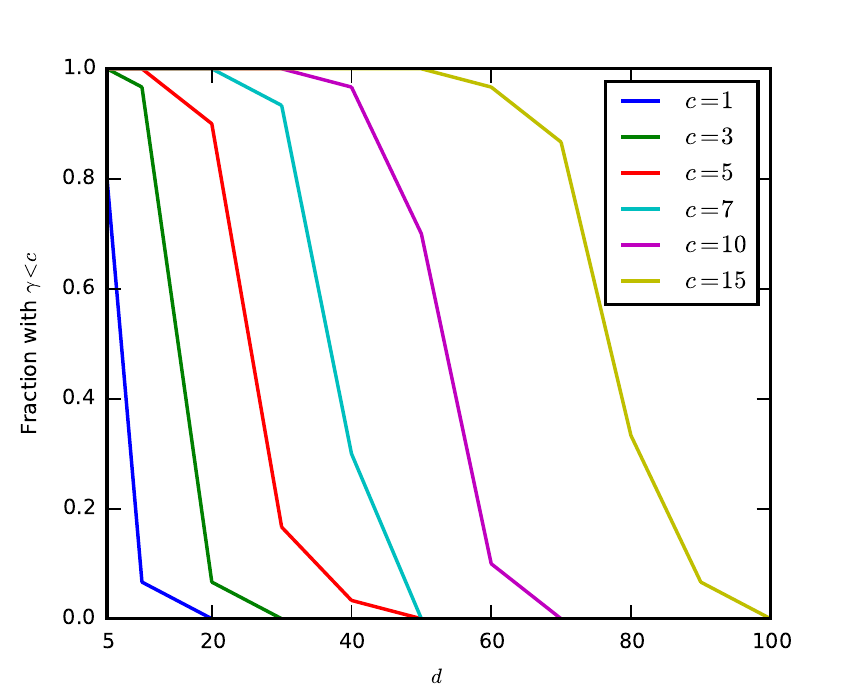}
  \caption{Testing the condition of Assumption 1 of
    \citet{Ravikumar2011} in Eq.~(\ref{eq:assumption1}) on real gene
    expression data showing the fraction of random subsets of $d$
    genes that fulfil the requirement and various relaxations.
    The condition~(\ref{eq:assumption1}) requires $\gamma < 1$,
    but the figure shows results also for larger $\gamma$ cutoffs.}
  \label{fig:tcga_assumption1}
\end{figure}

Fig.~\ref{fig:tcga_assumption1} shows the fraction of covariances
derived from random subsets of $d$ genes that satisfy the Assumption 1
of \citet{Ravikumar2011} ($c=1$) as well as the fraction of values
below more relaxed bounds.  The figure shows that the assumption is
reliably satisfied only for very small $d$ while for $d \ge 20$, the
assumption is essentially never satisfied.  Based on the results of
Fig.~\ref{fig:latent_assumption1} it is likely that glasso results
will degrade significantly by for $\gamma > 10$ and beyond which are
very common for large networks.

We further studied how accurately glasso can recover the graphical structures when the data were generated using the precision matrices described above. We used a similar thresholding with a cut-off value of $0.1$ in order to first form sparse precision matrices for a random subset of genes with given dimension. These matrices were then inverted to obtain covariance matrices. We checked that the resulting matrices were positive definite and then used them to sample multivariate normal data with zero mean with different sample sizes.

\begin{figure}[htb]
  \centering
  \includegraphics[width=\columnwidth]{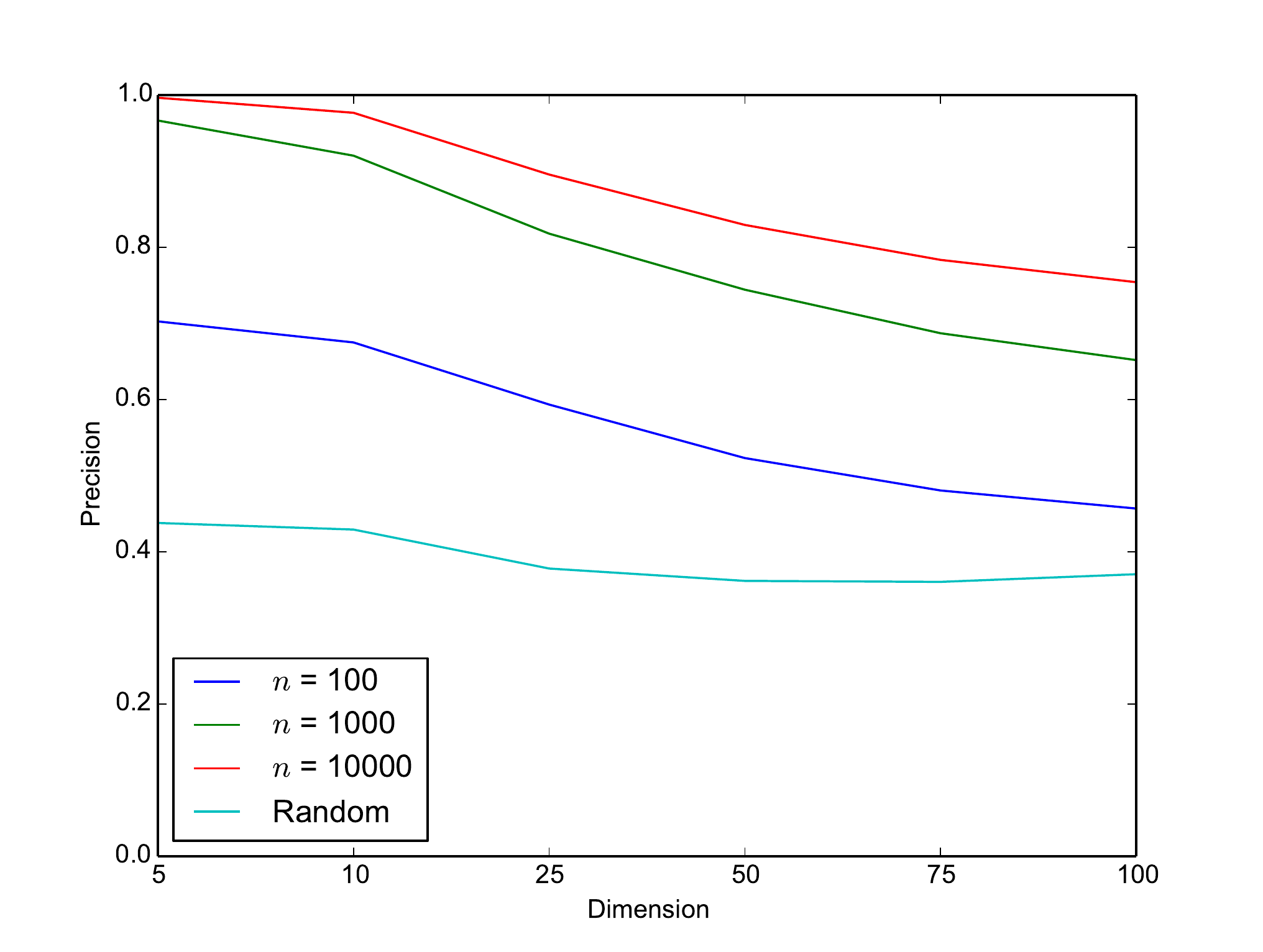}
  \caption{Average precisions for glasso with different
    dimensions and sample sizes of the real gene expression data,
    higher values are better.  The precision obtained by random
    guessing is also illustrated.}
  \label{fig:tcga_precision}
\end{figure}

The obtained data sets were centred and scaled before computing the sample covariance which was used as input to the glasso algorithm. The regularisation parameter was chosen with the aid of the ground truth graph, so that the the graph identified by glasso would contain as many edges as there were in the real graph. Results are shown in Figure \ref{fig:tcga_precision}.
The results show that glasso performance decreases as the network
size increases and is approaching that of random guessing for the
largest networks considered here.

\begin{figure}[htb]
  \centering
  \includegraphics[width=\columnwidth]{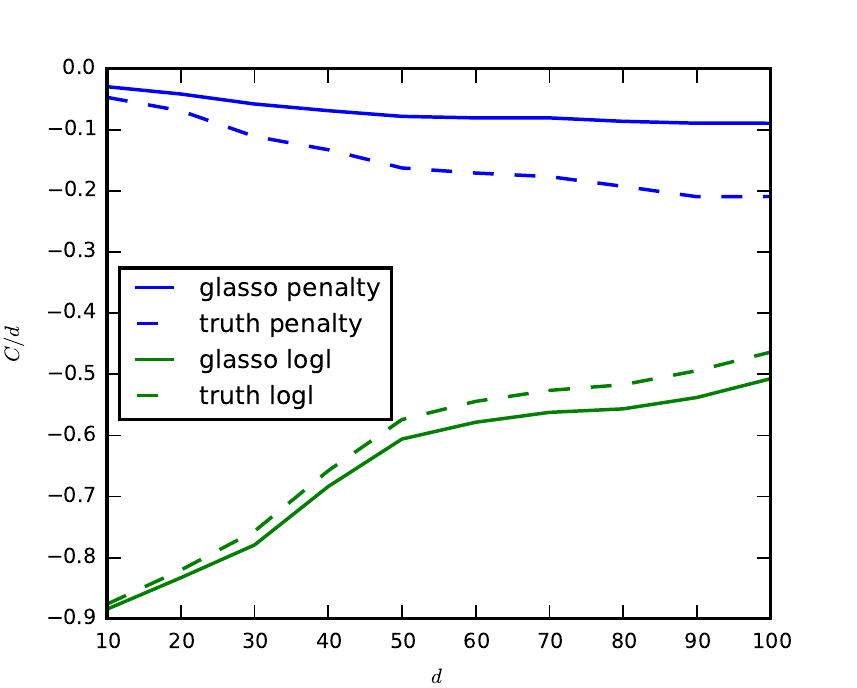}
  \caption{Average contributions of the different terms of the glasso
    objective function (\ref{eq:glassoObj}) on real gene expression
    data over random subsets of $d$ genes. The values are shown for
    the $\ell_1$ penalty term as well as the unnormalised
    log-likelihood, divided by $d$ to make them comparable.  Solid
    lines show the values for glasso result while dashed lines show
    the result for ground truth.}
  \label{fig:tcga_objective}
\end{figure}

Fig.~\ref{fig:tcga_objective} shows the contributions of different
parts of the glasso objective function (\ref{eq:glassoObj}) as a
function of the number of genes $d$.  The regularisation parameter
$\lambda$ of glasso was tuned to return a solution with the same number
of edges as in the true solution.  We used the glasso implementation of
scikit-learn \citep{scikit-learn}, which ignores the diagonal terms
of $\vOmega$ when computing the penalty.  The figure shows clearly how the
penalty term for the true solution increases superlinearly as a
function of $d$.  (A linear increase would correspond to a horizontal
line.)  The result is even more striking given that the optimal
$\lambda$ decreases slightly as $d$ increases.  The penalty
contribution for glasso solution increases much more slowly.  The
excess loss in log-likelihood from glasso solution increases as $d$
increases, but this is compensated by a larger saving in the penalty.
Together these suggest that glasso solutions are likely to remain further
away from ground truth as $d$ increases.

\section{DISCUSSION}

The class of latent variable like models presented in
Sec.~\ref{sec:latent-variable-like} is an interesting example of
models that have a very clear sparse structure, which all
$\ell_1$-penalisation-based methods seem unable to recover even in the
limit of infinite data.  This class complements the previously considered examples
of models where glasso is inconsistent including the ``two
neighbouring triangles'' model of \citet{Meinshausen2008} and the
star graph of \citet{Ravikumar2011}, the latter of which can be seen
as a simple special case of our example.

An important question arising from our investigation is how significant the discovered limitation to inferring sparse covariance matrices is in practice, i.e. how common are the latent variable like structures in real
data sets.  Given the popularity and success of linear models in
diverse applications it seems plausible such structures could often exist in
real data sets, either as an intrinsic property or as a result of some human
intervention, e.g.\ through inclusion of partly redundant variables.

The gene expression data set is a natural example of an application where graphical
model structure learning has been considered.  The original glasso
paper~\citep{Friedman2008} contained an example on learning gene
networks, although from proteomics data.  Other authors
\citep[e.g.][]{Ma2007} have applied Gaussian graphical models
and even glasso \citep[e.g.][]{Menendez2010} to gene network
inference from expression data.  Our experiments on the TCGA gene
expression data suggest that in such applications it is advisable to consider the
conditions for the consistency of $\ell_1$ penalised methods very
carefully when planning to apply those.

Previous publications presenting new methods for sparse precision
matrix have typically tested the method on synthetic examples where
the true precision matrix is specified to contain mostly small values.
Specifying the precision matrix provides a convenient way to generate test
cases as the sparsity pattern can be defined very naturally through it.  At the
same time, this excludes any models that have an ill-conditioned
covariance.  As shown by our example, such ill-conditioned covariances
arise very naturally from model structures that are plausible from the application perspective.

Ultimately, our results suggest that users of the numerous $\ell_1$ penalised methods should
be much more careful about checking whether the conditions of
consistency for precision matrix estimation
are likely to be fulfilled in the application area of interest.

\subsubsection*{Acknowledgements}

This work was supported by the Academy of Finland [259440 to A.H.,
251170 to J.C.] and the European Research Council [239784 to J.C.].


\begin{thebibliography}{20}
\providecommand{\natexlab}[1]{#1}
\providecommand{\url}[1]{\texttt{#1}}
\expandafter\ifx\csname urlstyle\endcsname\relax
  \providecommand{\doi}[1]{doi: #1}\else
  \providecommand{\doi}{doi: \begingroup \urlstyle{rm}\Url}\fi

\bibitem[Banerjee et~al.(2008)Banerjee, El~Ghaoui, and
  d'Aspremont]{Banerjee2008}
O.~Banerjee, L.~El~Ghaoui, and A.~d'Aspremont.
\newblock Model selection through sparse maximum likelihood estimation for
  multivariate {Gauss}ian or binary data.
\newblock \emph{Journal of Machine Learning Research}, 9:\penalty0 485--516,
  June 2008.

\bibitem[Cai et~al.(2011)Cai, Liu, and Luo]{Cai2011}
T.~Cai, W.~Liu, and X.~Luo.
\newblock A constrained $\ell_1$ minimization approach to sparse precision
  matrix estimation.
\newblock \emph{Journal of the American Statistical Association}, 106\penalty0
  (494):\penalty0 594--607, Jun 2011.

\bibitem[{Cancer Genome Atlas Network}(2012)]{TCGA2012}
{Cancer Genome Atlas Network}.
\newblock Comprehensive molecular portraits of human breast tumours.
\newblock \emph{Nature}, 490\penalty0 (7418):\penalty0 61--70, Oct 2012.

\bibitem[Friedman et~al.(2008)Friedman, Hastie, and Tibshirani]{Friedman2008}
J.~Friedman, T.~Hastie, and R.~Tibshirani.
\newblock Sparse inverse covariance estimation with the graphical lasso.
\newblock \emph{Biostatistics}, 9\penalty0 (3):\penalty0 432--441, Jul 2008.

\bibitem[Hsieh et~al.(2014)Hsieh, Sustik, Dhillon, and Ravikumar]{Hsieh2014}
C.~Hsieh, M.~A. Sustik, I.~S. Dhillon, and P.~D. Ravikumar.
\newblock {QUIC:} quadratic approximation for sparse inverse covariance
  estimation.
\newblock \emph{Journal of Machine Learning Research}, 15\penalty0
  (1):\penalty0 2911--2947, 2014.

\bibitem[Lauritzen(1996)]{LAURITZEN1996}
S.~Lauritzen.
\newblock \emph{Graphical Models}.
\newblock Clarendon Press, 1996.
\newblock ISBN 9780191591228.

\bibitem[Lepp{\"a}-aho et~al.(2016)Lepp{\"a}-aho, Pensar, Roos, and
  Corander]{FMPL}
J.~Lepp{\"a}-aho, J.~Pensar, T.~Roos, and J.~Corander.
\newblock Learning {G}aussian graphical models with fractional marginal
  pseudo-likelihood.
\newblock \emph{arXiv:1602.07863}, 2016.

\bibitem[Liu and Luo(2015)]{Liu2015}
W.~Liu and X.~Luo.
\newblock Fast and adaptive sparse precision matrix estimation in high
  dimensions.
\newblock \emph{Journal of Multivariate Analysis}, 135:\penalty0 153 -- 162,
  2015.

\bibitem[Lu and Shiou(2002)]{Lu2002}
T.-T. Lu and S.-H. Shiou.
\newblock Inverses of $2\times 2$~{b}lock matrices.
\newblock \emph{Computers \& Mathematics with Applications}, 43\penalty0
  (1-2):\penalty0 119–129, Jan 2002.

\bibitem[Ma et~al.(2007)Ma, Gong, and Bohnert]{Ma2007}
S.~Ma, Q.~Gong, and H.~J. Bohnert.
\newblock An {Arabidopsis} gene network based on the graphical {Gauss}ian
  model.
\newblock \emph{Genome Res}, 17\penalty0 (11):\penalty0 1614--1625, Nov 2007.

\bibitem[Meinshausen(2008)]{Meinshausen2008}
N.~Meinshausen.
\newblock A note on the {L}asso for {Gauss}ian graphical model selection.
\newblock \emph{Statistics \& Probability Letters}, 78\penalty0 (7):\penalty0
  880--884, May 2008.

\bibitem[Meinshausen and B\"{u}hlmann(2006)]{Meinshausen2006}
N.~Meinshausen and P.~B\"{u}hlmann.
\newblock High-dimensional graphs and variable selection with the lasso.
\newblock \emph{The Annals of Statistics}, 34\penalty0 (3):\penalty0
  1436--1462, Jun 2006.

\bibitem[Men{\'{e}}ndez et~al.(2010)Men{\'{e}}ndez, Kourmpetis, {ter Braak},
  and {van Eeuwijk}]{Menendez2010}
P.~Men{\'{e}}ndez, Y.~A.~I. Kourmpetis, C.~J.~F. {ter Braak}, and F.~A. {van
  Eeuwijk}.
\newblock Gene regulatory networks from multifactorial perturbations using
  {G}raphical {L}asso: application to the {DREAM4} challenge.
\newblock \emph{PLoS One}, 5\penalty0 (12):\penalty0 e14147, 2010.

\bibitem[Pedregosa et~al.(2011)Pedregosa, Varoquaux, Gramfort, Michel, Thirion,
  Grisel, Blondel, Prettenhofer, Weiss, Dubourg, Vanderplas, Passos,
  Cournapeau, Brucher, Perrot, and Duchesnay]{scikit-learn}
F.~Pedregosa, G.~Varoquaux, A.~Gramfort, V.~Michel, B.~Thirion, O.~Grisel,
  M.~Blondel, P.~Prettenhofer, R.~Weiss, V.~Dubourg, J.~Vanderplas, A.~Passos,
  D.~Cournapeau, M.~Brucher, M.~Perrot, and E.~Duchesnay.
\newblock Scikit-learn: Machine learning in {P}ython.
\newblock \emph{Journal of Machine Learning Research}, 12:\penalty0 2825--2830,
  2011.

\bibitem[Peng et~al.(2009)Peng, Wang, Zhou, and Zhu]{Peng09}
J.~Peng, P.~Wang, N.~Zhou, and J.~Zhu.
\newblock Partial correlation estimation by joint sparse regression models.
\newblock \emph{Journal of the American Statistical Association}, 104\penalty0
  (486):\penalty0 735--746, 2009.

\bibitem[Ravikumar et~al.(2011)Ravikumar, Wainwright, Raskutti, and
  Yu]{Ravikumar2011}
P.~Ravikumar, M.~J. Wainwright, G.~Raskutti, and B.~Yu.
\newblock High-dimensional covariance estimation by minimizing
  $\ell_1$-penalized log-determinant divergence.
\newblock \emph{Electronic Journal of Statistics}, 5:\penalty0 935--980, 2011.

\bibitem[Tibshirani(1996)]{Tibshirani1996}
R.~Tibshirani.
\newblock Regression shrinkage and selection via the lasso.
\newblock \emph{Journal of the Royal Statistical Society, Series B},
  58:\penalty0 267--288, 1996.

\bibitem[Whittaker(1990)]{WHITTAKER1990}
J.~Whittaker.
\newblock \emph{Graphical Models in Applied Multivariate Statistics}.
\newblock John Wiley \& Sons, 1990.

\bibitem[Yuan and Lin(2007)]{Yuan07}
M.~Yuan and Y.~Lin.
\newblock Model selection and estimation in the {Gauss}ian graphical model.
\newblock \emph{Biometrika}, 94\penalty0 (1):\penalty0 19--35, 2007.

\bibitem[Zhao and Yu(2006)]{Zhao2006}
P.~Zhao and B.~Yu.
\newblock On model selection consistency of lasso.
\newblock \emph{Journal of Machine Learning Research}, 7:\penalty0 2541--2563,
  2006.

\end{thebibliography}

\end{document}